\documentclass[twoside,11pt]{article}

\usepackage{amsmath}
\usepackage{amssymb}
\usepackage{parskip}
\usepackage{a4wide}
\usepackage{mathtools}
\usepackage{listings}
\usepackage{tikz}
\usepackage{pgfplots}
\usepackage[english]{babel}
\usepackage{graphicx}
\usepackage{subcaption}

\usepackage[utf8]{inputenc} 
\usepackage{hyperref}       
\usepackage{url}            
\usepackage{booktabs}       
\usepackage{amsfonts}       
\usepackage{nicefrac}       
\usepackage{microtype}      
\usepackage{multirow}
\usepackage{algorithm}
\usepackage{algorithmic}

\pgfplotsset{compat=newest}

\usepackage{natbib}

\newcommand{\tblref}[1]{Table~\ref{#1}}
\newcommand{\Tblref}[1]{Table~\ref{#1}}
\newcommand{\figref}[1]{Figure~\ref{#1}}

\newcommand{\algref}[1]{Algorithm~\ref{#1}}
\newcommand{\subrefpar}[1]{(\subref{#1})}

\newcommand{\domain}[1]{#1}

\DeclareMathOperator*{\maximize}{maximize}

\DeclareMathOperator*{\argmax}{argmax}
\newcommand{\degree}{\ensuremath{{}^\circ}}

\newcommand{\iter}[1]{^{(#1)}}
\newcommand{\h}{\mathbf{w}}
\newcommand{\x}{\mathbf{x}}
\newcommand{\numiter}{M}
\newcommand{\itvar}{k}
\newcommand{\numclass}{K}
\newcommand{\classvar}{c}

\newcommand{\hlinetop}{\toprule}
\newcommand{\hlinemid}{\midrule}
\newcommand{\hlinebot}{\bottomrule}
\newcommand{\colsep}{2mm}
\newcommand{\leftsep}{0mm}
\newcommand{\subheadersep}{.4ex}
\newcommand{\methodsep}{.8ex}
\makeatletter
\newcommand{\tablesize}{\@setfontsize\small\@viiipt\@ixpt}
\newcommand{\tablestdsize}{\@setfontsize\small\@viipt\@ixpt}
\makeatother
\newcommand{\inlinestd}[1]{{\tablestdsize$\pm$#1}}
\newcommand{\withstd}[1]{\raisebox{0.5mm}{\tablestdsize$\pm$#1}}
\newcommand{\stddevs}{\vspace*{-0.5mm}}
\newcommand{\best}[1]{{\fontseries{b}\selectfont #1}}

\newcommand{\NAME}{Ad-REM}

\title{Domain Adaptation with Randomized Expectation Maximization}
\author{%
   Twan van Laarhoven
   tvanlaarhoven@cs.ru.nl\\
   Elena Marchiori
   elenam@cs.ru.nl\\
   Radboud University, Postbus 9010, 6500GL Nijmegen, The Netherlands
  }

\begin{document}


\maketitle

\begin{abstract}%
Domain adaptation (DA) is the task of classifying an unlabeled dataset (target) using a labeled dataset (source) from a related domain. 
The majority of successful DA methods try to directly match the distributions of the source and target data by transforming the feature space.
Despite their success, state of the art methods based on this approach are either involved or unable to directly scale to data with many features.  
%
This article shows that domain adaptation can be successfully performed by using a very simple randomized expectation maximization (EM) method.  
We consider two instances of the method, which involve logistic regression and  support vector machine, respectively. 
The underlying assumption of the proposed method is the existence of a good single linear classifier for both source and target domain.
The potential limitations of this assumption are alleviated by the flexibility of the method, which can directly incorporate deep features extracted from a pre-trained deep neural network. 
The resulting algorithm is strikingly easy to implement and apply.
We test its performance on $36$ real-life adaptation tasks over text and image data with diverse characteristics.
The method achieves state-of-the-art results, competitive with those of involved end-to-end deep transfer-learning methods.

Source code is available  at \url{http://github.com/twanvl/adrem}.
\end{abstract} 

\section{Introduction}
\label{sec:intro}


Domain adaptation (DA) addresses the problem of building a good predictor for a target domain using labeled training data from a related source domain and target unlabeled training data. A typical example in visual object recognition involves two different datasets  consisting of images taken under different cameras or conditions: for instance, one dataset consists of images taken at home  with a digital camera while another dataset contains images taken in a controlled environment  with  studio lightning conditions.



A natural formulation of the DA problem is: finding a model performing well on both source and target data. Without any prior knowledge (that is, with a uniform prior on each model),  the source classifier is the best expected  choice. In order to do `better' on the target domain, one needs to incorporate some other desirable property.   

A desirable property adopted by most DA methods is the `small domain discrepancy'. The popularity of this property is grounded on theoretical results on domain adaptation, which showed that the domain discrepancy between source and target contributes to the target error \citep{ben2007analysis,ben2010theory}. 

Most current state-of-the-art methods reduce the discrepancy between source and target distributions using feature transformations.  Simple methods based on this approach include \cite{Fernando2013SA,gong2013connecting,sun2016return}. These methods have superquadratic complexity in the number of features, which is why they are not directly applied to high dimensional data. To overcome this problem feature selection on the source data is used, which reduces the predictive information for the target domain. For instance this phenomenon can happen in natural language processing, where different genres often use very different vocabulary to described similar concepts. As classical example of this situation, consider product reviews retrieved from Amazon.com and the task of classifying them as positive or negative (sentiment polarity analysis): reviews of electronic type of  products and reviews of dvd products use different vocabularies. Hence adaptation between these two domains (electronic and dvd) is rather challenging \cite{blitzer2007biographies}.

Recently, end-to-end DA methods based on deep neural networks (DNN's), like \cite{ganin2016domain,sener2016learning,long2016unsupervised},  have been shown to perform very well on visual adaptation tasks. Training a DNN may require a large train data \citep{sener2016learning} as well as the use of target labeled data to tune parameters \citep{long2016unsupervised}. Furthermore DNN's for domain adaptation  can be sensitive to (hyper-)parameters of the learning procedure \citep{ganin2016domain}. To address these issues, state of the art end-to-end DA methods fine tune deep neural networks which were trained on related very large data. For instance, in DA for visual object recognition, deep neural networks pre-trained on the Imagenet dataset are used, such as VGG \citep{Simonyan14vgg}. 

Pre-trained DNN's have been used also to enhance the performance of DA methods based on shallow neural network architectures, like linear support vector machines (SVM's) \citep[e.g.][]{sun2016return}. In this setting, a pre-trained DNN is employed in a pre-processing step to generate a non-linear representation of the input space based on deep features. However, as we will also show in our experimental analysis,  shallow methods based on domain discrepancy have a limited positive effect when applied to deep features.

Although impressive results on domain adaptation have been achieved in the past decade, drawbacks of the current state of the art motivate our investigation.  Our goal is to develop a DA method that is very simple, scalable to many features, and competitive with end-to-end DNN methods when used with deep features.

To this aim we re-visit an old friend of the machine learning community: the Expectation Maximization (EM) approach for labeled and unlabeled data \citep[see e.g.][]{mclachlan1975iterative}. The idea of this approach is to view the labels on unlabeled samples as missing data.  \citet{seeger2000learning} provide an excellent review of this approach.


EM has been used in the past for semi-supervised learning  \cite{amini2003semi,nigam2006semi,grandvalet2006entropy,smola2005kernel} and self-training  \cite{ghahramani1994supervised,li2008self,tan2009adapting, bruzzone2010domain}. The so-called “hard” EM approach use provisional target label, while the “soft” EM one incorporates label confidences when fitting the model on the next iteration \cite{margolis2011literature}. 
In self-training, the labeled data is used to train an initial model, which is then used to guess the labels or label probabilities of the unlabeled data. On the next round, the unlabeled data with provisional labels  (or label probabilities) are incorporated to train a new model. This procedure is iterated for a fixed number of times, or until convergence. Methods based on this approach differ in the way target samples are added and used. Some methods add only the samples with high label confidence at each iteration, e.g. \cite{bruzzone2010domain}, while others use all the target data on each round, e.g. \cite{li2008self}.


These methods have been significantly outperformed by more recent DA methods. A core reason of their sub-optimal performance is that EM suffers from the problem that poor label or label probability estimates on one round can lead to fast convergence to bad pseudo labels close to those estimated in the first E-step \citep{seeger2000learning,margolis2011literature}. 

%


We propose to overcome the problem of fast convergence to bad pseudo labels by performing the M-step only on a random sub-sample of the target data.
The size of the sample is then increased at each iteration until the entire target data is considered. 

The only desirable property we embed in the method is `class balance' to avoid degenerate solutions where all data have the same class label. We use a controlled random sampling which generates classes of equal size.  

The resulting method, called {\NAME} ({\it Ad}aptation with {\it R}andomized {\it EM}), uses the source labeled data to train an initial model (we consider a linear SVM or a logistic regression model), which is then used to guess the labels of the unlabeled target data. On the next round, a class-balanced random sub-sample the unlabeled target data with provisional labels is used to train a new model. The procedure is iterated for a fixed number of times. To reduce the variance due to the randomized procedure, we ensemble pseudo-labelings results over multiple runs.

{\NAME} is strikingly simple to implement and apply.  It does not involve learning a transformation of the feature space, as in current state of the art (shallow) DA methods. Thus it is directly applicable to adaptation tasks which involve many variables. We show that on $12$ adaptation tasks in natural language processing with many features \cite{blitzer2007biographies}, where shallow methods like CORAL \cite{sun2016return} cannot be directly applied, a neat improvement is achieved by {\NAME}. 

{\NAME} is extremely simple in comparison with end-to-end deep learning domain adaptation methods, which need large training data or rely on pre-trained deep models for weight initialization, and sometimes employ a small set of labelled target data for tuning their hyper-parameters  \citep{bousmalis2016domain,long2016unsupervised}.  {\NAME} can be directly used with deep features from pre-trained DNN's.  We perform an extensive experimental analysis with three publicly available pre-trained DNN architectures.
We show that using deep features, {\NAME} consistently achieves impressive results on object recognition adaptation tasks. 

In particular, using ResNet deep features, {\NAME} achieves 96.7\% average accuracy over the 12 adaptation tasks of the Office-Caltech 10 dataset, and 87.0\% average accuracy over the 6 adaptation tasks of  the Office 31 dataset \citep{saenko2010adapting}. On these datasets, current state of the art shallow method SA \citep{Fernando2013SA} and CORAL \citep{sun2016return}  do not improve  over the baseline SVM model without adaptation, while end-to-end deep DA methods achieve inferior performance.



Overall, results  of experiments on $36$ real-life adaptation tasks in visual object recognition and natural language  demonstrate the state-of-the-art performance of {\NAME}.    
 
In summary our contributions are: (1) a new DA method based on randomized EM; (2) two strikingly easy DA algorithms based on this method. These algoithms are (3) scalable to data with many features; (4) flexible to the use of deep features from pre-trained deep neural networks;
(5) competitive or better than involved end-to-end deep learning DA algorithms.  

These contributions provide renewed support to EM, as an easy and effective approach to perform domain adaptation.


\section{Related work}
\label{sec:related}

For an in-depth description of the EM approach and methods for learning with labeled and unlabeled data we refer the reader to \citet{seeger2000learning}. Algorithms based on the EM-approach are also well described by \citet[section 3.1.4]{margolis2011literature}. In particular, a self-training semi-supervised SVM algorithm has been introduced by \citet{li2008self}. This EM-based algorithm iteratively re-trains a SVM classifier using an unlabeled dataset. At each iteration the unlabeled dataset is used for training with pseudo-labels computed from the model at the previous iteration.  In the DA context, the use of the entire unlabelled (target) dataset to update the model at each iteration is ineffective, since it favors fast convergence to the source classifier, as we will show in next section.

The literature on DA is vast, for which we refer to survey papers like \cite{margolis2011literature,pan2010survey,patel2015visual}.  Here we limit our  description to few methods based on popular approaches.

{\em Domain discrepancy reduction.} The majority of the algorithms for DA try to reduce the discrepancy between source and target distributions using a data transformation. Popular and simple methods in this class include \citet{Fernando2013SA,gong2013connecting,sun2016return}.  For instance,  CORrelation ALignment (CORAL) finds a linear transformation that minimizes the distance between the covariance of source and target.  CORAL \citep{sun2016return} uses a specific mapping and is not scalable for high number of features because it needs to compute and invert a covariance matrix.
Subspace Alignment (SA) computes a linear map that minimizes the Frobenius norm  of  the  difference  between the source and target domains, which are represented by subspaces described by eigenvectors \citep{Fernando2013SA}. \citet{gong2012geodesic} proposed a Geodesic Flow Kernel (GFK) which models domain shift by integrating an infinite number of subspaces that characterize changes in geometric and statistical properties from
the source to the target domain. 
These  methods learn a feature transformation and cannot be directly applied to high dimensional input data, since they have quadratic complexity in the number of features.


 {\em Importance weighting}. Importance-weighting methods assign a weight to each source instance in such a way as to make the reweighed version of the source distribution as similar to the target distribution as possible \citep{shimodaira2000improving}. Despite their theoretic appeal, importance-weighting approaches generally do not to perform very well when there is little ``overlap'' between the source and target domain.  

{\em Feature level adaptation}. Feature level domain-adaptation methods either extend the source data and the target data with additional features that are similar in both domains \citep{blitzer2006domain}. The more recent Feature Level Domain Adaptation (FLDA) method by \citet{kouw2016feature} models the dependence between the two domains by means of a feature-level transfer model that is trained to describe the transfer from source to target domain. FLDA assigns a data-dependent weight to each feature representing how informative it is in the target domain. To to do it uses information on differences in feature presence between the source and the target domain.

{\em Self training}. The approach underlying self-training methods has been explained in the introduction.   One of the first algorithms for domain adaptation based on this approach was introduced by \citet{bruzzone2010domain}, which extended the formulation of support vector machines to domain adaptation. The method progressively  adjusts an SVM classifier trained on the source data toward the target domain by replacing source samples with target samples having high confidence prediction. This way to update the classifier - choosing the most confidently labeled target samples according to the previous iteration’s estimation -  although robust, yields a minor change the current belief, that is, the resulting classifier remains close to the source one.  This is the reason why this method has been significantly outperformed by more recent DA algorithms which try to minimize domain discrepancy. Other methods based on this approach include the method by \citet{habrard2013boosting} which optimizes both the source classification error and margin constraints over the unlabeled target instances, and includes a regularization term to favor the reduction of the divergence between source and target distributions, and  the methods by \citet{germain2016new} which derive DA bounds for the weighted majority vote framework that are used to infer principled DA algorithms. 

{\em End-to-end deep neural networks}. Recently end-to-end DA methods based on deep neural networks have been shown to achieve impressive performance \citep{ganin2016domain,sener2016learning,long2016unsupervised}. For instance, DLID \citep{chopra2013dlid} is an end-to-end deep adaptation method which learns multiple intermediate  representations  along  an  interpolating path between the source and target domains.  Deep Transfer Network (DTN) \cite{zhang2015deep} employs a deep neural network to model and match both the domains marginal and conditional distributions. Residual Transfer Networks (RTN) \cite{long2016unsupervised} performs feature adaptation by matching the feature distributions of multiple layers across domains. Deep-CORAL \citep{sun2016deep} aligns correlations of layer activations in deep neural networks.  Recently, adversarial learning has become a popular approach for domain adaptation. For instance, \citet{tzeng2015simultaneous} proposed adding a binary domain classifier to discriminate domain labels and a domain confusion loss to enforce its prediction to become close to a uniform distribution over binary labels.  ReverseGrad \citep{ganin2015unsupervised,ganin2016domain} enforces the domains to be indistinguishable by reversing the gradients of the the loss of the domain classifier.  Joint Adaptation Networks (JAN) \citep{long2016deep} 
aligns the joint distributions of multiple domain-specific layers across domains by means of a joint maximum mean discrepancy measure which is optimized using an adversarial learning procedure. Disadvantages of end-to-end DA methods based on deep neural networks are the need of large train data \citep{sener2016learning}, the use of target labels to tune parameters \citep{long2016unsupervised} and their sensitivity to (hyper-)parameters of the learning procedure \citep{ganin2016domain}.

Compared to recent state-of-the-art-algorithms, our method  does not perform feature transformation or feature learning for adaptation, which is more effective in the presence of many features.  This is also  advantageous when deep features from pre-trained models are used, since it may be ineffective to match source and target distribution in such deep-feature space.  The superiority of our method in this setting is demonstrated by the results of our experiments. In particular, on visual object recognition, state of the art shallow methods like CORAL \citep{sun2016return}  and SA \citep{Fernando2013SA} achieve little or no gain when used with deep features extracted from pre-trained neural networks, while our method outperforms involved end-to-end DNN algorithms like
JAN \citep{long2016deep}
\section{Domain Adaptation with Randomized Expectation Maximization}
\label{sec:method}

We follow \citet{margolis2011literature} to explain the DA setting we use. We have a labeled (source) dataset $S=\{(\x^S_i,y^S_i)\}_{i=1}^{|S|}$ and an unlabeled (target) dataset $T=\{\x^T_i\}_{i=1}^{|T|}$.  We want to maximize the observations of labeled source data $S$ and unlabeled target data $T$, with the target labels as hidden variables. The EM objective is:

$$\maximize_{\h} \sum_{i\in S}  \log(p(x^S_i,y^S_i \mid \h)) +  \sum_{i\in T}  \log E_ {y{^T_i} \mid \h} \{ p(x^T_i \mid y^T_i,\h)\}.$$

The EM algorithm applied to this problem leads to a version of self-labeling with “soft” labels, where on iteration $\itvar$, the E- and M-step are as follows:
\begin{description}
\item [E:] Given a model $p(x,y\mid \h_\itvar)$, compute $p(y^T_i = c \mid  x^T_i,\h_\itvar)$ for all $x^T_i$ and all class labels $c$.
\item [M:] $\h_{\itvar+1} = \argmax
\sum_{i\in S}  \log(p(x^S_i,y^S_i\mid \h)) + \lambda \sum_{i\in T}  \log E_ {y{^T_i}\mid  \h_\itvar} \{ p(x^T_i\mid  y^T_i,\h)\}$.
\end{description}

Typically, $\h$ is initialized on $S$, e.g. by maximizing $\sum_{i\in S}  \log(p(x^S_i,y^S_i \mid  \h))$.

In our domain adaptation context, optimization is also with respect to the labels of the target data, and the objective of hard EM becomes:

\begin{equation}
  \label{eq:sl}
\maximize_{\h, y^T_i} \sum_{i\in S}  \log(p(x^S_i,y^S_i\mid \h)) +  \sum_{i\in T}  \log p(x^T_i,y^T_i\mid \h)\}.
\end{equation}

This corresponds to a version of self-training with hard labels on $T$: 

\begin{description}
\item [E:] Given a model $p(x,y\mid\h_\itvar)$, compute $y^T_i = \argmax_c  p(y^T_i = c\mid  x^T_i,\h_\itvar)$ for all $x^T_i$.
\item  [M:]  $\h_{\itvar+1} = \argmax
\sum_{i\in S}  \log(p(x^S_i,y^S_i\mid \h)) + \sum_{i\in T}  \log  p(x^T_i \mid  y^T_i,\h)$.
\end{description}


The hard EM objective differs from the soft EM one in the second term computed over target domain examples: hard EM tries to maximize $p(x,y \mid \h)$,  whereas soft EM tries to maximize only $p(x \mid \h)$.

The exact implementation of the maximization step depends on the model, we use logistic regression and SVM.

%



{\em Logistic Regression (LR)}: The regularized log-likelihood for the logistic regression model is:
 \begin{equation}
  \label{eq:lr}
L_{LR}(\h) = \sum_{i\in S} \ell(\h \cdot x^S_i, y^S_i) - \log[1+\exp(\h \cdot x^S_i)]+ \sum_{i\in T} \ell(\h \cdot x^T_i,y^T_i) - \log[1+\exp(\h \cdot x^T_i)] - \lambda ||w||^2,
\end{equation}
where $\ell(z,y) = zy$.

The M-step of the self-training procedure for the LR model maximizes the function \ref{eq:lr}.

%
%


{\em Support Vector Machine}: Although SVM is not a probabilistic model,  \citet{grandvalet2006probabilistic} showed  that  the  hinge  loss  can  be  interpreted  as  the neg-log-likelihood of a semi-parametric model of posterior probabilities, while \citet{franc2011support} showed how the SVM can be viewed as a maximum likelihood estimate of a class of semi-parametric models. Therefore it is also `legal' to use the SVM loss in our EM setting for DA:

\begin{equation}
  \label{eq:svm}
  L_{SVM}(\mathbf{\h}) = C \sum_{i\in S} \ell(\h \cdot x^S_i,y^S_i) +
                       C \sum_{i\in T} \ell(\h \cdot x^T_i,y^T_i) +
                       \|\h\|_2^2,
\end{equation}
where $\ell(z,y) = \max(0,1-yz)$ is the hinge loss and $C>0$ is a hyper-parameter.

The M-step of the self-training procedure for the SVM model maximizes the function \ref{eq:svm}.

The above described EM approach has been developed and used in semi-supervised learning, where labeled and unlabeled samples are drawn from the same distribution.
This is not necessarily the case when doing domain adaptation, since the unlabeled target data comes from a related yet different domain then the labeled source data.
A consequence is that EM style methods do too little exploration. The resulting classifier stays close to one trained on the source data, but this tends to be a poor local optimum when the target is different from the source. 

We describe in the next sub-sections our proposal to overcome this problem: class-balanced random sub-sample selection. We use SVM to illutrate our approach. A similar reasoning applies for the LR model.

\subsection{Class-balanced Randomized EM}

To overcome the limitations of the EM to quickly converge to potentially bad local optima, we propose to add randomization.
%
%
Our approach is based on two observations.

\newcommand{\Tother}{\hat{T}}
\newcommand{\hother}{\hat{\h}}

First of all, suppose that we have found an optimum $\hother$ of $L$ for another sample $\Tother$ from the target domain.
Using the framework of empirical risk minimization it follows that $\hother$ also generalizes to new samples from the target domain.
In particular, it will give a good classification of $T$.

Once we have a classification of the target samples, we can use it as $y^T$. Then \eqref{eq:svm} reduces to a standard supervised SVM loss, which is easy to optimize using off-the-self libraries.

This leaves the problem of finding an optimal classifier $\hother$ for a sample $\Tother$.

Our second observation is that the smaller the size of the target dataset, the closer the problem becomes to a standard support vector machine on the source data. In the extreme case, when $|T|=0$, we can easily optimize $L$ exactly. Conversely, if $|\Tother|$ is close to $|T|$ we expect that $\hother$ is a better classifier for $T$.

We therefore propose an iterative sampling strategy, where we gradually increase the size of the sample of target data on which \eqref{eq:svm} is optimized.
Since we can not draw new samples from the target domain, we instead draw samples from the given target dataset $T$.

%
%

Another problem with the EM-based optimization of $L_\text{SVM}$ is that even the global optimum might not actually give a good classification of the target domain. In particular, if there is a large margin between the two domains, then separating all of the target domain into a single class often gives a low loss.
This issue becomes especially apparent in high dimensions, where the amount of empty space between data points becomes larger.

A way to avoid this problem is to enforce class balance, that is, that each class occurs roughly equally often among the labels $y^T$.
Here we enforce class balance through a sampling strategy: at each iteration of our EM method we use a controlled random selection of the sub-sample of the target data with current pseudo-labels to enforce class balance. This corresponds to the optimization of a modified SVM loss which includes class balance by assigning equal importance to all classes, rather than to all target domain samples.
Denote by $T_{\classvar}$ the set of points that is assigned to class $\classvar$ in $y^T$, i.e. $T_{\classvar}=\{i \mid y^T_i = \classvar\}$. And denote by $\numclass$ the number of classes.
Then we define the balanced DA SVM loss to be
\begin{equation}
  \label{eq:btsvm}
L_\text{balanced}(\h)
            = C \sum_{i=1}^{|S|} \ell(\h \cdot \x^S_i, y^S_i) + 
              \frac{C}{\numclass} \sum_{\classvar=1}^\numclass \frac{|T|}{|T_\classvar|} \sum_{i \in T_\classvar} \ell(\h \cdot \x^T_i, y^T_i) +
              \|\h\|_2^2.
\end{equation}

The resulting algorithm, which we call {\NAME} (Adaptation with Randomized EM), is shown in \algref{alg:method}.

The above balanced DA SVM loss is close to the transductive support vector machine (TSVM) one. The only difference is the way class-balanced is incorporated in the optimization problem: directly in the DA SVM loss and as a constraint in TSVM. No algorithm is known to efficiently find a globally optimal solution of the TSVM optimization problem  \citep{joachims2006transductive}.
Thus {\NAME} can be viewed as a randomized method to solve  the  TSVM optimization problem.  

%
%
%
%

\begin{figure*}
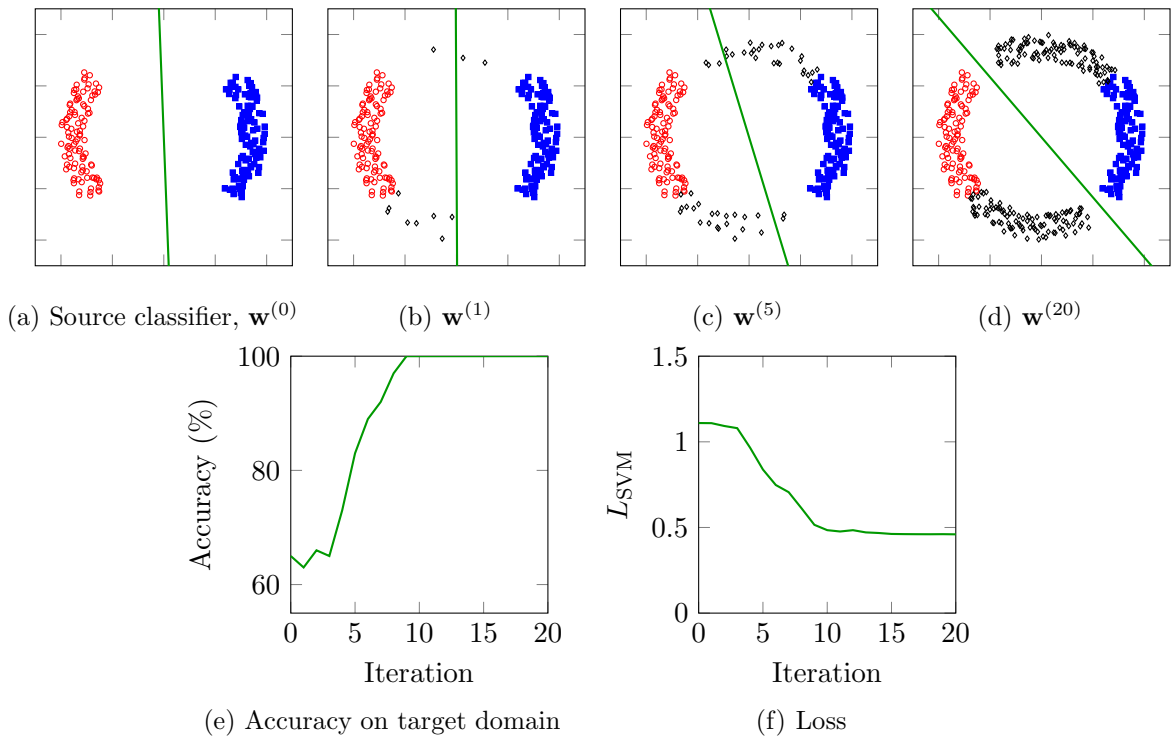

  \centering
  \begin{subfigure}{0.25\linewidth}
    \centering

    \caption{Source classifier, $\h\iter{0}$}
    \label{fig:toy:first}
  \end{subfigure}%
  \begin{subfigure}{0.25\linewidth}
    \centering
    %
    \caption{$\h\iter{1}$}
  \end{subfigure}%
  \begin{subfigure}{0.25\linewidth}
    \centering
    %
    \caption{$\h\iter{5}$}
  \end{subfigure}%
  \begin{subfigure}{0.25\linewidth}
    \centering
    %
    \caption{$\h\iter{20}$}
    \label{fig:toy:last}
  \end{subfigure}
  \begin{subfigure}{0.35\linewidth}
    %
    \caption{Accuracy on target domain}
    \label{fig:toy:accuracy}
  \end{subfigure}
  \begin{subfigure}{0.35\linewidth}
    %
    \caption{Loss}
    \label{fig:toy:loss}
  \end{subfigure}

  \caption{
    A toy example of domain adaptation.
    The source data span an arc of $80\degree$ for each class, and the target data is rotated $80\degree$ compared to the source, leaving a gap of $20\degree$ with the source data of the other class.
    \subrefpar{fig:toy:first}\dots\subrefpar{fig:toy:last} show the trained classifier over the iterations of {\NAME}, as larger samples are selected for training the SVM.
    \subrefpar{fig:toy:loss} shows the SVM loss $L_\text{SVM}(\h)$ over the iterations.
  }
  \label{fig:toy}
\end{figure*}

In \figref{fig:toy} we illustrate the execution of {\NAME} on a toy example.
It can be clearly seen that as the algorithm progresses the loss decreases, and the number of correctly classified samples in the target domain increases.
In this example the correct classifier clearly has a large margin, so it can be found by minimizing $L_\text{SVM}$.

\begin{figure*}
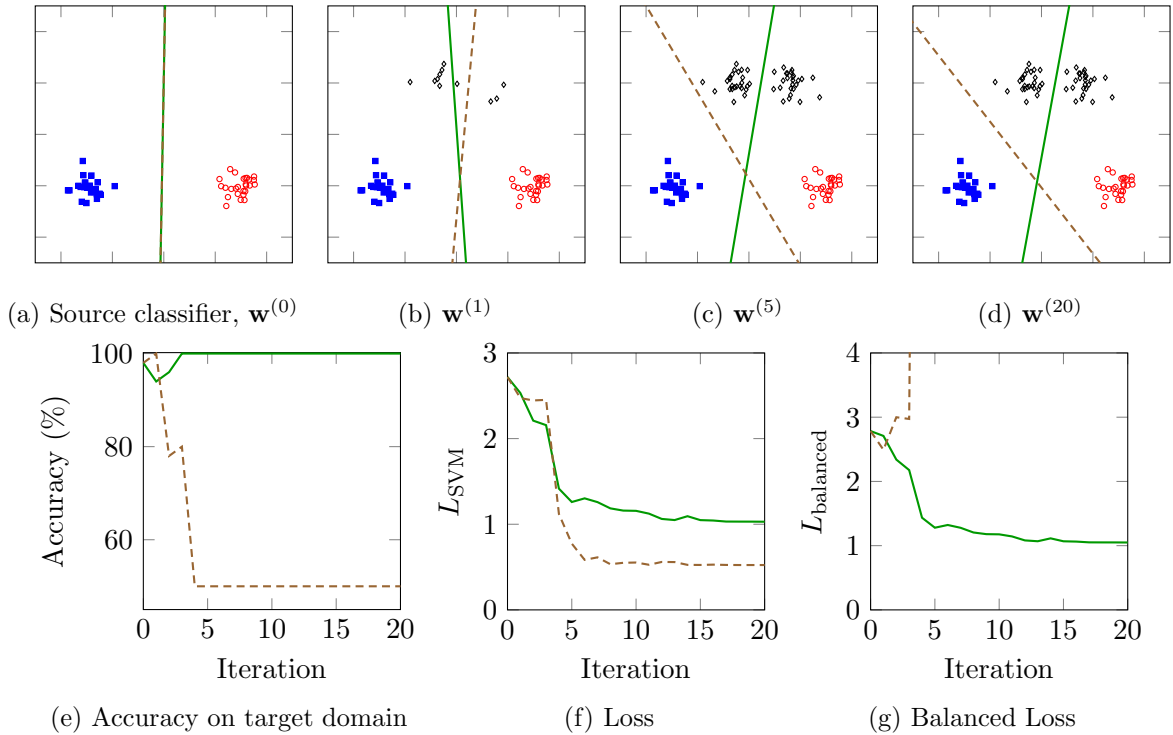

  \centering
  \begin{subfigure}{0.25\linewidth}
    \centering

    \caption{Source classifier, $\h\iter{0}$}
    \label{fig:toy-balance:first}
  \end{subfigure}%
  \begin{subfigure}{0.25\linewidth}
    \centering
    %
    \caption{$\h\iter{1}$}
  \end{subfigure}%
  \begin{subfigure}{0.25\linewidth}
    \centering
    %
    \caption{$\h\iter{5}$}
  \end{subfigure}%
  \begin{subfigure}{0.25\linewidth}
    \centering
    %
    \caption{$\h\iter{20}$}
    \label{fig:toy-balance:last}
  \end{subfigure}
  \begin{subfigure}{0.34\linewidth}
    %
    \caption{Accuracy on target domain}
    \label{fig:toy-balance:accuracy}
  \end{subfigure}%
  \begin{subfigure}{0.31\linewidth}
    %
    \caption{Loss}
    \label{fig:toy-balance:loss}
  \end{subfigure}%
  \begin{subfigure}{0.31\linewidth}
    %
    \caption{Balanced Loss}
    \label{fig:toy-balance:balanced-loss}
  \end{subfigure}
  \caption{
    A toy example showing the need for balancing.
    The dashed brown classifier is trained without balancing, resulting in a better SVM loss, but a much lower accuracy.
  }
  \label{fig:toy-balance}
\end{figure*}

In \figref{fig:toy-balance} we illustrate the need of class balance. It can be seen that although without class balance the loss of SVM improves during the iterations, the target accuracy decreases.

%

Because {\NAME} uses random samples of the data, the results can depend on the exact samples chosen.
To reduce the variance in the predictions we  simply run the method multiple times, and take a majority vote of the resulting labelings.
In all experiments we have used the ensemble-{\NAME} method, which is shown in \algref{alg:ensemble}.

\begin{algorithm}
  \caption{The single-{\NAME} method}
  \label{alg:method}
  \begin{algorithmic}
    \REQUIRE Source dataset $S=(\x^S,y^S)$; Target dataset $T=\x^T$; Regularization parameter $C$; number of iterations $\numiter$.
    \STATE $\h\iter{0} \gets \text{train SVM on } S$
    \STATE $y\iter{0} \gets \text{predict labels with } \h\iter{0} \text{ on } T$
    \FOR{$\itvar \gets 1$ \TO $\numiter$}
      \STATE $n_\itvar \gets \frac{\itvar}{\numiter}|T|$
      \STATE $B\iter{\itvar} \gets \text{draw balanced subset of size } n_i \text{ from } (\x^T, y\iter{\itvar-1})$
      \STATE $\h\iter{\itvar} \gets \text{train SVM on } \text{concat}(S,B\iter{\itvar})$ \; \; \%\text{ {\bf M-step}}
      \STATE $y\iter{\itvar} \gets \text{predict labels with } \h\iter{\itvar} \text{ on } T$ \; \; \, \%\text{ {\bf E-step}}
    \ENDFOR
    \RETURN $y\iter{\numiter}$
  \end{algorithmic}
\end{algorithm}

\begin{algorithm}
  \caption{The ensemble-{\NAME} method}
  \label{alg:ensemble}
  \begin{algorithmic}
    \REQUIRE Source dataset $S=(x^S,y^S)$; Target dataset $T=\x^T$; Regularization parameter $C$; number of iterations $M$; number of ensemble iterations $m$.
    \FOR{$j \gets 1$ \TO $m$}
    \STATE $z_j \gets \text{single-\NAME}(S,T,C,M)$
    \ENDFOR
    \RETURN MajorityVote$(z_1,z_2,\dotsc,z_m)$
  \end{algorithmic}
\end{algorithm}


\def\sz{5cm}
\pgfplotsset{
  toy/.style={%
    width=\sz,
    height=\sz,
    enlargelimits=false,xmin=-5,xmax=5,ymin=-5,ymax=5,
    xticklabel=\empty,
    yticklabel=\empty,
    scatter/classes={
      -1={mark=square*,blue,mark size=1},
       1={mark=o,red,mark size=1},
      -2={mark=square*,blue!30!white,mark size=1},
       2={mark=o,red!30!white,mark size=1},
       0={mark=diamond,black,mark size=1}
    }
  },
  toy2/.style={toy},
  dataset/.style={
    scatter,
    only marks,
    scatter src=explicit
  }
}
\tikzset{
  classifier/.style={
    solid,green!60!black,thick
  },
  classifier2/.style={
    densely dashed,brown!80!black,thick
  }
}
\pgfplotsset{
  select coords between index/.style 2 args={
    x filter/.code={
        \ifnum\coordindex<#1\def\pgfmathresult{}\fi
        \ifnum\coordindex>#2\def\pgfmathresult{}\fi
    }
  }
}

%

%
%

\section{Experimental analysis}
  
We perform extensive experiments on $36$ adaptation tasks from real-life text and image benchmark datasets of diverse characteristics: with high number of features, relatively small sample size, larger number of classes and large scale data. Furthermore, on two visual domain adaptation datasets, we perform extensive experiments with deep features generated from existing pre-trained deep neural network classifiers.
%
%

In all experiments we assume target instances to be unlabeled.
%
%
In {\NAME} we use an ensemble with 11 repetitions (we chose a prime number to prevent ties).
For all runs we use 20 iterations.
The parameter $C$ is chosen with internal three-fold cross-validation on the source domain.

We use linear SVM and Logistic Regression (LR) as base classifiers, implemented by liblinear \citep{Fan2008liblinear}.
Linear SVM and LR are our source hypothesis baselines that do not use domain adaptation. 
In our experiments with more than two classes, we used one-vs-all linear SVM and multi-class logistic regression. 


We compare with popular state-of-the-art shallow DA methods,
 Subspace Alignment (SA) \citep{Fernando2013SA},
Feature Level Domain Adaptation (FLDA) \citep{kouw2016feature}, and Correlation Alignment (CORAL) \citep{sun2016return}.
The latter is the dominant state-of-the-art approach employing feature transformation.

Parameters for these shallow baselines are chosen with cross validation on the source dataset. 


We assess all algorithms in a fully transductive setup where all unlabeled target instances are used during training for predicting their labels.
We use labeled instances of the first domain as the source and unlabeled instances of the second domain as the target. We evaluate the accuracy on the target domain as the fraction of correctly labeled target instances. 


We run experiments with source code of the shallow DA methods (except on the Office 31 dataset where we use results reported by \citet{sun2016return} which were obtained using the same evaluation protocol we use for {\NAME}).  

We also assess the performance of {\NAME} and the shallow baselines on features extracted with a deep neural network. Specifically, we use the ResNet 50 architecture \cite{he2016deep} that was pre-trained on ImageNet. This network is available through Keras \citep{chollet2015keras}.
We rescale the images to $224\times224$ pixels, and pass these through the network. We then use the output of the nonlinearities on the last hidden layer as features.

\subsection{Results}

In the results reported in the sequel, we use {A$\to$B} to indicate the adaptation task with A as source dataset and B as target one. 

When results of a method are missing from a table it means either the method could not be run on such data (kept running for days) or the corresponding paper did not contain results of that method on the considered data.


\newcommand{\itparagraph}[1]{\textit{#1}}
\itparagraph{Amazon sentiment dataset}
This dataset, introduced by \citet{blitzer2007biographies}, has {many features} (over 47000), which are word unigram and bigram counts. It  involves $4$ domains, \domain{Books} (B), \domain{Dvd} (D), \domain{Electronics} (E) and \domain{Kitchen} (K), each with 1000 positive and 1000 negative examples obtained from the dichotomized 5-star rating.
The considered shallow baselines have quadratic complexity in the number of features, which is why they can't be used in high-dimensional settings, such as the Amazon dataset.  Therefore, these baselines have to use feature selection:  \citet{gong2013connecting} used feature selection to reduce the data set to 400 features. We conduct experiments with this dataset and  validation protocol as in \citet{sun2016return}: random subsamples of the source (1600 samples) and target (400 samples) data and standardized features. The experiment is repeated 20 times.

Feature selection might remove features relevant to the target. Keeping all features is possible in our method, which avoids this risk. Therefore we also report results of {\NAME} under the same experimental protocol but with {\it all} features.
In this case we can not standardize the data, because that would destroy the sparsity, instead we normalize by dividing each feature by its standard deviation.
To test the stability of the method we have repeated this experiment 10 times.
\Tblref{tbl:accuracy-amazon2-subset} reports the mean and standard deviation of the accuracy, which shows that using all features improves accuracy.
\itparagraph{Office-Caltech 10 object recognition dataset} This dataset \citep{gong2012geodesic} consists of $10$ classes of images from an office environment in 4 image domains: Webcam (W), DSLR (D), Amazon (A), and Caltech256 (C), with $958, 295, 157, 1123$ instances, respectively. 
The dataset uses 800 SURF features, which we preprocess by dividing by the instance-wise mean followed by standardizing.
We follow the standard protocol \citep{gong2012geodesic,Fernando2013SA,sun2016return}, and use 20 labeled samples per class from the source domain (except for the DSLR source domain, for which we use 8 samples per class).
We repeated these experiment 20 times, and report the mean and standard deviation of accuracy in \tblref{tbl:accuracy-office-caltech-standard}.


On this dataset CORAL, FLDA and {\NAME} achieve comparable performance when SURF features are used. The improvement over the baseline is very small, and for some transfer tasks such as {A$\to$C}, {\NAME} actually gives worse results than the baseline.

When using deep features, {\NAME} achieves best performance with a substantial increase in accuracy over no adaptation, while CORAL and FLDA do not improve over no adaptation.  Using ResNet 50 deep features, {\NAME} achieves 12\%  gain in the accuracy on the {C$\to$W}  transfer task (from 86.0\% to 98.1\%). Overall, results indicate that the source classifier achieves excellent performance when applied to deep features. These results confirm that  deep models trained on the very large ImageNet dataset generate powerful domain invariant features.

We have also performed experiments using the full source domain as training data, the results of this experiment are reported in \tblref{tbl:accuracy-office-caltech-full}.
These results are very similar to the results with the standard protocol. With SURF features there is a noticeable increase in accuracy compared to the standard protocol, because more source data is available. For the deep features the differences are much smaller, because the accuracy was already high.




\itparagraph{Office 31 dataset} We next perform object recognition adaptation with a  {\it larger number of classes} and deep features. We use the standard Office dataset 31 \citep{saenko2010adapting} which contains 31 classes (the 10 from the Office-Caltech 10 plus 21 additional ones)  in 3 domains: Webcam (W), DSLR (D), and Amazon (A). Office-31 has a total of 4110 instances, with a maximum of 2478 per domain over 31 classes.
%
We run experiments using all labeled source and unlabeled target data.


On this dataset CORAL achieves best performance, slightly better than that of {\NAME} with LR as base classifier, only when using the DECAF-fc7 deep features by \citet{tommasi2014testbed}. When using deep features  generated from the ResNet 50 architecture, {\NAME} outperforms all other methods, and obtains excellent performance, with a significant improvement over no adaptation.  The gain in the accuracy is rather large when performing adaptation over hard transfer tasks. For instance about 15\% gain is achieved on  {A$\to$W} (from 73.8\% to 89.4\%) and about 12\% on {D$\to$A} (from 60.3\% to 72.6\%).

\itparagraph{Cross Dataset Testbed} Finally, we also consider a {larger scale evaluation} using the Cross Dataset Testbed \citep{tommasi2014testbed}, using  deep features obtained with DECAF. The dataset contains $40$ classes from 3 domains: $3847$ images for the domain Caltech256 (C), $4000$ images for Imagenet (I), and $2626$ images for SUN (S).  Results of these experiments are shown in \tblref{tbl:accuracy-cdt}.
Also on this dataset {\NAME} obtains best results, and improves by a large margin over no adaptation. 
Previous papers have used standardization of the features.

We found it beneficial to increase sparsity by rectifying the inputs before normalization, that is, set negative values to 0 ($\max(0,x)$);
and to then normalize by dividing by the standard deviation only.
We see improved performance with rectified features compared to the original ones, likely because rectification was also used during the training of the neural network.
\subsection{Comparison with end-to-end deep methods}\label{sec:deep}
For a direct comparison with state-of-the-art end-to-end DNN methods, we compare results of {\NAME}  on the Office 31 dataset with  published results of deep transfer methods based on ResNet:
Deep Domain Confusion (DDC) \citep{tzeng2014deep},
Deep Adaptation Network (DAN) \citep{long2015learning},
Residual Transfer Networks (RTN) \cite{long2016unsupervised},
Reverse Gradient (RevGrad) \cite{ganin2015unsupervised,ganin2016domain}, and
Joint Adaptation Networks (JAN) \cite{long2016deep}.
All comparing methods fine-tune the ResNet50 architecture pre-trained on ImageNet. RTN and JAN are not fully unsupervised and need a few labeled target data for hyperparameter optimization.
These results are taken from \cite{long2016deep}, and reported in \tblref{tbl:accuracy-office}.

\subsection{Discussion}\label{sec:discussion}
On almost linearly separable classification tasks, like sentiment polarity classification with text data, {\NAME} and the other shallow methods achieve comparable results when applied to a subset of pre-selected $400$ features. However,  {\NAME}  can profit from the direct use of all $47000$ features of this dataset, and achieve improved target accuracy, while the other shallow algorithms do not terminate or run out of memory when all features are used.

On harder classification tasks, like visual object detection, target accuracy performance does not differ substantially across  shallow methods.
In particular, when the performance of the source classifier on the target is low, like on the  of the Office-Caltech 10 dataset with SURF features, performance of {\NAME} remains close to that of the source classifier, and is sometimes worse. This happens because a good starting point is needed for the algorithm to converge to a good local optimum.

Using deep features from pre-trained deep neural network models is highly beneficial for {\NAME}. In this case, the source classifier is good also for the target, and {\NAME} profits from this, especially on  {A$\to$D}, {A$\to$W} and {C$\to$W}, where it gains about 10\% accuracy. On the other hand, the gain in accuracy obtained using other shallow DA methods is relatively small. 


Overall, results with deep features on the Office-Caltech 10 and Office 31 datasets show that the source classifier already achieves good performance when applied to deep features.  These results confirm that deep models trained on the very large ImageNet dataset generate powerful domain invariant features. 

\subsection{Sensitivity analysis}\label{sec:sensitivity}


In this paragraph we investigate the sensitivity of {\NAME} to its parameters.

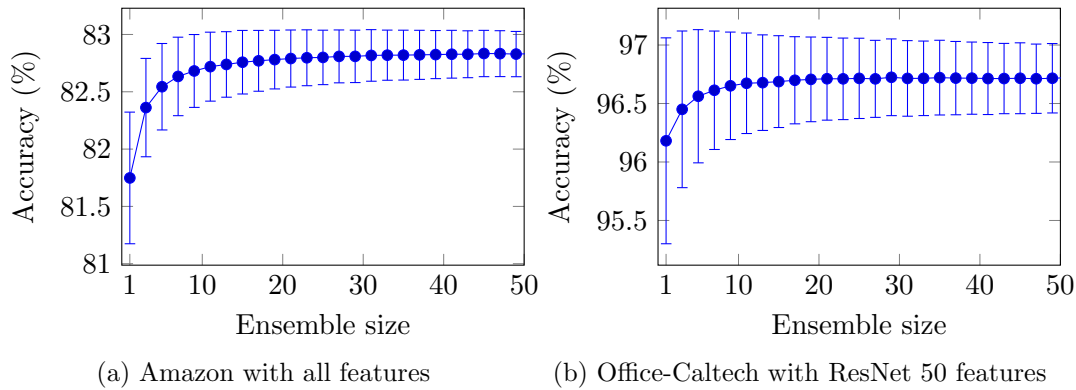
\begin{figure*}
  \centering
  \begin{subfigure}{0.45\linewidth}
    \begin{tikzpicture}
      \begin{axis}
        [width=0.99\linewidth
        ,height=5cm
        ,xmin=0, xmax=50
        ,xtick={1,10,20,30,40,50}
        ,xlabel={Ensemble size}, ylabel={Accuracy (\%)}]
        \addplot+[error bars/.cd, y dir=both, y explicit] table[x=size,y expr=\thisrow{mean_acc}*100, y error expr=\thisrow{std_acc}*100]{
size  mean_acc std_acc
1  0.817484 0.005742
3  0.823621 0.004290
5  0.825441 0.003771
7  0.826339 0.003424
9  0.826816 0.003178
11  0.827192 0.003008
13  0.827385 0.002858
15  0.827580 0.002773
17  0.827701 0.002653
19  0.827807 0.002558
21  0.827900 0.002489
23  0.827967 0.002438
25  0.827998 0.002375
27  0.828087 0.002307
29  0.828089 0.002286
31  0.828171 0.002250
33  0.828193 0.002187
35  0.828198 0.002175
37  0.828223 0.002141
39  0.828244 0.002097
41  0.828276 0.002085
43  0.828274 0.002045
45  0.828339 0.002020
47  0.828326 0.002002
49  0.828289 0.001975
51  0.828327 0.001939
53  0.828319 0.001947
55  0.828336 0.001908
57  0.828359 0.001889
59  0.828357 0.001878
61  0.828353 0.001856
63  0.828355 0.001847
65  0.828354 0.001822
67  0.828358 0.001807
69  0.828358 0.001800
71  0.828382 0.001800
73  0.828407 0.001768
75  0.828392 0.001765
77  0.828391 0.001754
79  0.828389 0.001730
81  0.828404 0.001713
83  0.828387 0.001697
85  0.828388 0.001701
87  0.828386 0.001689
89  0.828393 0.001673
91  0.828394 0.001659
93  0.828387 0.001658
95  0.828402 0.001629
97  0.828405 0.001640
99  0.828384 0.001617
};
      \end{axis}
    \end{tikzpicture}
    \caption{Amazon with all features}
  \end{subfigure}
  \begin{subfigure}{0.45\linewidth}
    \begin{tikzpicture}
      \begin{axis}
        [width=0.99\linewidth
        ,height=5cm
        ,xmin=0, xmax=50
        ,xtick={1,10,20,30,40,50}
        ,xlabel={Ensemble size}, ylabel={Accuracy (\%)}]
        \addplot+[error bars/.cd, y dir=both, y explicit] table[x=size,y expr=\thisrow{mean_acc}*100, y error expr=\thisrow{std_acc}*100] {
size  mean_acc std_acc
1  0.961811 0.008802
3  0.964502 0.006697
5  0.965621 0.005694
7  0.966132 0.005063
9  0.966513 0.004594
11  0.966730 0.004297
13  0.966783 0.004086
15  0.966872 0.003925
17  0.966987 0.003725
19  0.967064 0.003620
21  0.967117 0.003540
23  0.967109 0.003486
25  0.967150 0.003432
27  0.967108 0.003293
29  0.967242 0.003280
31  0.967145 0.003231
33  0.967153 0.003186
35  0.967213 0.003194
37  0.967171 0.003134
39  0.967168 0.003086
41  0.967134 0.003078
43  0.967134 0.003005
45  0.967159 0.003030
47  0.967119 0.002962
49  0.967159 0.002967
51  0.967210 0.002932
53  0.967143 0.002922
55  0.967175 0.002916
57  0.967173 0.002895
59  0.967132 0.002845
61  0.967167 0.002810
63  0.967214 0.002834
65  0.967139 0.002762
67  0.967170 0.002796
69  0.967147 0.002773
71  0.967161 0.002746
73  0.967168 0.002746
75  0.967160 0.002663
77  0.967165 0.002680
79  0.967122 0.002613
81  0.967110 0.002695
83  0.967127 0.002598
85  0.967186 0.002702
87  0.967129 0.002556
89  0.967176 0.002598
91  0.967073 0.002583
93  0.967149 0.002575
95  0.967207 0.002587
97  0.967183 0.002544
99  0.967136 0.002531
};
      \end{axis}
    \end{tikzpicture}
    \caption{Office-Caltech with ResNet 50 features}
  \end{subfigure}
  \caption{
    Accuracy as a function of ensemble size $m$. Average over all domain pairs, and 2000 repetitions. Error bars indicate standard deviation.
  }
  \label{fig:ensemble}
\end{figure*}

 
First, in \figref{fig:ensemble} we plot the test set accuracy as a function of the ensemble size $m$.
{\NAME} is a stochastic method, since it relies on bootstrap samples in each iterations.
Using an ensemble is a way of reducing this variance.
As expected, using a larger ensemble produces better results.
The variance over multiple runs also becomes smaller with a larger ensemble size.
For $m>10$ there are diminishing returns to further increasing the ensemble size.
%

\begin{figure*}
  \centering
  \begin{subfigure}{0.45\linewidth}
    \begin{tikzpicture}
      \begin{axis}
        [width=0.99\linewidth
        ,height=5cm
        ,xmin=1, xmax=50
        ,xtick={1,10,20,30,40,50}
        ,xlabel={Number of iterations}, ylabel={Accuracy (\%)}]
        \addplot table[x=num_iterations,y expr=\thisrow{mean_acc}*100] {
num_iterations  mean_acc
1  0.758785 
5  0.799498 
10  0.810488 
15  0.814532 
20  0.818089 
25  0.817941 
30  0.819623 
35  0.821472 
40  0.820592 
45  0.821287 
50  0.821335 
55  0.821623 
60  0.822805 
65  0.822196 
70  0.822641 
75  0.822680 
80  0.822765 
85  0.822455 
90  0.822925 
95  0.823428 
100  0.823520 
};
      \end{axis}
    \end{tikzpicture}
    \caption{Amazon with all features}
  \end{subfigure}
  \begin{subfigure}{0.45\linewidth}
    \begin{tikzpicture}
      \begin{axis}
        [width=0.99\linewidth
        ,height=5cm
        ,xmin=1, xmax=50
        ,xtick={1,10,20,30,40,50}
        ,xlabel={Number of iterations}, ylabel={Accuracy (\%)}]
        \addplot table[x=num_iterations,y expr=\thisrow{mean_acc}*100] {
num_iterations  mean_acc
1  0.929862 
5  0.951673 
10  0.957627 
15  0.960259 
20  0.961742 
25  0.962264 
30  0.963013 
35  0.963120 
40  0.963762 
45  0.963635 
50  0.963830 
55  0.964048 
60  0.963964 
65  0.963865 
70  0.964185 
75  0.964073 
80  0.964351 
85  0.963990 
90  0.964051 
95  0.964077 
100  0.964123 
};
      \end{axis}
    \end{tikzpicture}
    \caption{Office-Caltech with ResNet 50 features}
  \end{subfigure}
  \caption{
    Accuracy as a function of number of iterations $M$.
    No ensemble is used.
  }
  \label{fig:num-iterations}
\end{figure*}
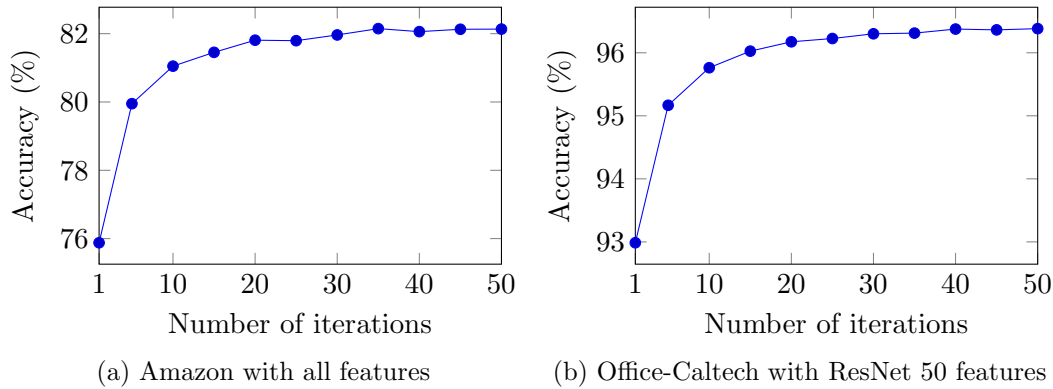

In \figref{fig:num-iterations} we vary the number of iterations $M$, without using an ensemble ($m=1$).
Again, more iterations produce better results, since the algorithm has more time to converge to good labels on the target domain.


\pgfplotscreateplotcyclelist{oversamples}{%
  red,            mark repeat=50, every mark/.append style={fill=.!80!red},  mark=*\\%
  blue,           mark repeat=50, every mark/.append style={fill=.!80!blue}, mark=square*\\%
  green!60!black, mark repeat=50, every mark/.append style={fill=.!80!black},mark=triangle*\\%
  black,          mark repeat=50, mark=star\\%
}
\pgfplotscreateplotcyclelist{oversamples-zoom}{%
  red,            mark repeat=5, every mark/.append style={fill=.!80!red},  mark=*\\%
  blue,           mark repeat=5, every mark/.append style={fill=.!80!blue}, mark=square*\\%
  green!60!black, mark repeat=5, every mark/.append style={fill=.!80!black},mark=triangle*\\%
  black,          mark repeat=5, mark=star\\%
}

\section{Conclusion}
 
We introduced {\NAME}, a strikingly easy and effective method for  DA. We showed that {\NAME}  can be viewed as a EM-based method with a controlled random sampling strategy, which enforces class balance on the target domain during the optimization procedure.

Results of our experimental analysis lead to the following interesting conclusions: (1) EM can be successfully exploited in domain adaptation;  (2) the direct combination of {\NAME} with deep learning features is highly beneficial and competitive with more involved end-to-end deep-transfer learning methods, notably on hard transfer tasks, such as $D \to A$ and $W\to A$ in the Office 31 dataset; (3) {\NAME}  sets a new state-of-the-art on various transfer tasks with image as well as text data.

There are (at least) two  limitations of our new approach for DA that remain to be investigated, which we summarize below. 
%

{\em Piecewise linearity}.  {\NAME} considers linear hypotheses generated by training SVM in the original feature space: the final majority vote model is a piecewise linear classifier.  Therefore, for almost linearly separable tasks, like the adaptation tasks of the Amazon dataset, its performance is excellent. However, this may be a limitation in case of highly non-separable classification tasks.  Results of our experiments showed that {\NAME } can directly profit from the used of deep features from pre-trained models in visual domain adaptation, with a significant increase in performance. It remains to be investigated whether it is even more beneficial to incorporate {\NAME } into an end-to-end deep learning architecture for DA or use a more poweful type of base classifier.

{\em Fixed number of iterations}. Our iterative EM-based procedure  terminates after a number of iterations which are determined by the input parameter $M$. We showed in our sensitivity analysis that {\NAME } is robust with respect to the choice of $M$.
However, there is no theoretical guarantee that the algorithm will have converged after $M$ iterations.
A natural extension of the method could be to let the algorithm proceed without sampling after the last iteration, as in the standard EM method.
To still optimize the class balanced loss function, one could use weighting instead of biased sampling in these extra iterations.
In practice the lack of convergence is not a problem, since there is already no guarantee of finding a global optimum.
However, the effect of alternative and possibly more effective controlled random sampling strategies with guaranteed convergence remain to be investigated.
\section*{Acknowledgements}
We would like to thank Baochen Sun for his prompt reply to our emails, in particular for providing interesting information about the application of CORAL.

This work has been partially funded by the Netherlands Organization for Scientific Research (NWO) within the EW TOP Compartiment 1 project 612.001.352.


%
\begin{table}[t]
  \caption{
    Accuracy on the Amazon sentiment dataset using the standard protocol of \citet{gong2013connecting,sun2016return}.
  }
  \label{tbl:accuracy-amazon2-subset}
  \tablesize
  \centering
\begin{tabular}{@{\hspace*{\leftsep}}ll@{\hspace*{\colsep}}c@{\hspace*{\colsep}}c@{\hspace*{\colsep}}c@{\hspace*{\colsep}}c@{\hspace*{\colsep}}c}
\hlinetop
& & K$\to$D & D$\to$B & B$\to$E & E$\to$K & avg\\
\hlinemid
\multicolumn{5}{l}{400 features}\\[\subheadersep]
& Source SVM & {73.3}\inlinestd{1.9} & {78.3}\inlinestd{2.3} & {75.6}\inlinestd{1.5} & {83.1}\inlinestd{1.8} & {77.6}\inlinestd{1.1}\\
& Source LR & {74.2}\inlinestd{1.6} & {78.9}\inlinestd{2.1} & {76.4}\inlinestd{2.2} & {84.0}\inlinestd{1.7} & {78.4}\inlinestd{1.1}\\[\methodsep]
& SA & {73.3}\inlinestd{1.9} & {78.3}\inlinestd{2.3} & {75.6}\inlinestd{1.5} & {83.1}\inlinestd{1.8} & {77.6}\inlinestd{1.1}\\
& FLDA-L & {73.5}\inlinestd{2.0} & {79.4}\inlinestd{2.1} & {75.6}\inlinestd{1.9} & {84.3}\inlinestd{2.0} & {78.2}\inlinestd{0.9}\\
& FLDA-Q & {74.5}\inlinestd{1.8} & {79.6}\inlinestd{1.8} & {77.2}\inlinestd{1.6} & {84.6}\inlinestd{1.8} & {79.0}\inlinestd{1.0}\\
& CORAL & {73.5}\inlinestd{1.8} & {78.3}\inlinestd{2.0} & {76.1}\inlinestd{1.7} & {83.1}\inlinestd{1.9} & {77.7}\inlinestd{0.8}\\
\end{tabular}
\end{table}

\begin{table*}[h]
 \caption{Accuracy on the Amazon dataset using the full source and target domain. Mean and standard deviation over 10 runs.}
  \label{tbl:accuracy-amazon-full}
  \tablesize
  \centering
\begin{tabular}{@{\hspace*{\leftsep}}ll@{\hspace*{\colsep}}c@{\hspace*{\colsep}}c@{\hspace*{\colsep}}c@{\hspace*{\colsep}}c@{\hspace*{\colsep}}c@{\hspace*{\colsep}}c@{\hspace*{\colsep}}c@{\hspace*{\colsep}}c@{\hspace*{\colsep}}c@{\hspace*{\colsep}}c@{\hspace*{\colsep}}c@{\hspace*{\colsep}}c@{\hspace*{\colsep}}c}
\hlinetop
& & B$\to$D & B$\to$E & B$\to$K & D$\to$B & D$\to$E & D$\to$K & E$\to$B & E$\to$D & E$\to$K & K$\to$B & K$\to$D & K$\to$E & avg\\
\hlinemid
\multicolumn{13}{l}{400 features}\\[\subheadersep]
& Source SVM & {77.5} & {76.4} & {78.2} & {78.2} & {75.3} & {78.2} & {73.8} & {73.1} & {83.2} & {74.3} & {73.6} & {82.7} & {77.0}\\
\stddevs&&\withstd{0.0}&\withstd{0.0}&\withstd{0.0}&\withstd{0.0}&\withstd{0.0}&\withstd{0.0}&\withstd{0.0}&\withstd{0.0}&\withstd{0.0}&\withstd{0.0}&\withstd{0.0}&\withstd{0.0}&\withstd{0.0}\\
& Source LR & {78.9} & {77.4} & {79.7} & {78.7} & {77.4} & {80.0} & {73.6} & {73.4} & {84.8} & {75.0} & {74.7} & {83.0} & {78.0}\\
\stddevs&&\withstd{0.0}&\withstd{0.0}&\withstd{0.0}&\withstd{0.0}&\withstd{0.0}&\withstd{0.0}&\withstd{0.0}&\withstd{0.0}&\withstd{0.0}&\withstd{0.0}&\withstd{0.0}&\withstd{0.0}&\withstd{0.0}\\[\methodsep]
& SA & {77.5} & {76.4} & {78.2} & {78.2} & {75.3} & {78.2} & {73.8} & {73.1} & {83.2} & {74.3} & {73.6} & {82.7} & {77.0}\\
\stddevs&&\withstd{0.0}&\withstd{0.0}&\withstd{0.0}&\withstd{0.0}&\withstd{0.0}&\withstd{0.0}&\withstd{0.0}&\withstd{0.0}&\withstd{0.0}&\withstd{0.0}&\withstd{0.0}&\withstd{0.0}&\withstd{0.0}\\
\end{tabular}
\end{table*}

\begin{table*}[ht]
  \caption{
    Average accuracy on the Office-Caltech 10 dataset using the standard protocol of \citet{gong2012geodesic,Fernando2013SA,sun2016return}.
    We report standard deviations where available.
  }
  \label{tbl:accuracy-office-caltech-standard}
  \tablesize
  \centering
\begin{tabular}{@{\hspace*{\leftsep}}ll@{\hspace*{\colsep}}c@{\hspace*{\colsep}}c@{\hspace*{\colsep}}c@{\hspace*{\colsep}}c@{\hspace*{\colsep}}c@{\hspace*{\colsep}}c@{\hspace*{\colsep}}c@{\hspace*{\colsep}}c@{\hspace*{\colsep}}c@{\hspace*{\colsep}}c@{\hspace*{\colsep}}c@{\hspace*{\colsep}}c@{\hspace*{\colsep}}c}
\hlinetop
& & A$\to$C & A$\to$D & A$\to$W & C$\to$A & C$\to$D & C$\to$W & D$\to$A & D$\to$C & D$\to$W & W$\to$A & W$\to$C & W$\to$D & avg\\
\hlinemid
\multicolumn{13}{l}{SURF features}\\[\subheadersep]
& Source SVM & {36.3} & {36.7} & {35.9} & {45.0} & {42.1} & {39.1} & {34.6} & {32.1} & {75.8} & {37.9} & {33.9} & {73.5} & {43.6}\\
\stddevs&&\withstd{1.6}&\withstd{3.3}&\withstd{2.5}&\withstd{2.2}&\withstd{3.3}&\withstd{3.8}&\withstd{1.0}&\withstd{0.9}&\withstd{2.4}&\withstd{0.5}&\withstd{1.2}&\withstd{3.4}&\withstd{1.0}\\
& Source LR & {36.3} & {36.7} & {34.9} & {44.9} & {41.7} & {38.8} & {34.5} & {31.8} & {76.2} & {38.0} & {34.3} & {75.3} & {43.6}\\
\stddevs&&\withstd{1.5}&\withstd{3.4}&\withstd{2.4}&\withstd{1.9}&\withstd{3.4}&\withstd{2.9}&\withstd{0.9}&\withstd{1.1}&\withstd{2.3}&\withstd{0.9}&\withstd{1.3}&\withstd{2.1}&\withstd{0.8}\\[\methodsep]
& SA & {43.0} & {37.6} & {37.1} & {47.3} & {42.2} & {38.3} & {38.1} & {33.7} & {79.2} & {37.3} & {33.4} & {78.2} & {45.4}\\
\stddevs&&\withstd{0.0}&\withstd{3.5}&\withstd{2.4}&\withstd{2.0}&\withstd{3.1}&\withstd{4.5}&\withstd{1.4}&\withstd{1.1}&\withstd{1.6}&\withstd{1.5}&\withstd{1.4}&\withstd{2.8}&\withstd{0.8}\\
\end{tabular}
\end{table*}

\begin{table*}[ht]
 \caption{Accuracy on the Office-Caltech 10 dataset, using the full source domain for training.}
 \label{tbl:accuracy-office-caltech-full}
 \tablesize
 \centering
\begin{tabular}{@{\hspace*{\leftsep}}ll@{\hspace*{\colsep}}c@{\hspace*{\colsep}}c@{\hspace*{\colsep}}c@{\hspace*{\colsep}}c@{\hspace*{\colsep}}c@{\hspace*{\colsep}}c@{\hspace*{\colsep}}c@{\hspace*{\colsep}}c@{\hspace*{\colsep}}c@{\hspace*{\colsep}}c@{\hspace*{\colsep}}c@{\hspace*{\colsep}}c@{\hspace*{\colsep}}c}
\hlinetop
& & A$\to$C & A$\to$D & A$\to$W & C$\to$A & C$\to$D & C$\to$W & D$\to$A & D$\to$C & D$\to$W & W$\to$A & W$\to$C & W$\to$D & avg\\
\hlinemid
\multicolumn{13}{l}{SURF features}\\[\subheadersep]
& Source SVM & {41.0} & {40.1} & {42.0} & {52.7} & {45.9} & {47.5} & {33.0} & {32.1} & {75.9} & {38.4} & {34.6} & {75.2} & {46.5}\\
& Source LR & {42.3} & {40.1} & {39.7} & {54.6} & {45.2} & {49.5} & {35.9} & {32.2} & {82.7} & {37.9} & {34.0} & {79.6} & {47.8}\\[\methodsep]
& SA & {37.4} & {36.3} & {39.0} & {44.9} & {39.5} & {41.0} & {32.9} & {34.3} & {65.1} & {34.4} & {31.0} & {62.4} & {41.5}\\
& FLDA-L & {41.5} & {45.9} & {42.0} & {49.5} & {48.4} & {44.1} & {31.7} & {34.1} & {75.6} & {35.3} & {33.8} & {72.6} & {46.2}\\
& FLDA-Q & {43.5} & {43.3} & {40.7} & {53.5} & {44.6} & {45.1} & {30.8} & {31.2} & {73.2} & {35.2} & {32.1} & {75.8} & {45.7}\\
& CORAL & {45.1} & {39.5} & {44.4} & {52.1} & {45.9} & {46.4} & {37.7} & {33.8} & {84.7} & {35.9} & {33.7} & {86.6} & {48.8}\\
\end{tabular}
\end{table*}

\begin{table}[ht]
  \caption{Accuracy on the Office 31 dataset.}
  \label{tbl:accuracy-office}
  \tablesize
  \centering
\begin{tabular}{@{\hspace*{\leftsep}}ll@{\hspace*{\colsep}}c@{\hspace*{\colsep}}c@{\hspace*{\colsep}}c@{\hspace*{\colsep}}c@{\hspace*{\colsep}}c@{\hspace*{\colsep}}c@{\hspace*{\colsep}}c}
\hlinetop
& & A$\to$D & A$\to$W & D$\to$A & D$\to$W & W$\to$A & W$\to$D & avg\\
\hlinemid
\multicolumn{7}{l}{DECAF-fc7 features}\\[\subheadersep]
& Source SVM & {47.6} & {46.0} & {46.6} & {85.4} & {41.5} & {90.8} & {59.6}\\
& Source LR & {53.0} & {50.2} & {44.1} & {85.1} & {42.0} & {89.4} & {60.6}\\[\methodsep]
& SA & {46.2} & {42.5} & {39.3} & {78.9} & {36.3} & {80.6} & {54.0}\\
& FLDA-Q & {54.2} & {50.5} & {40.4} & {80.3} & {38.4} & {86.5} & {58.4}\\
& CORAL & {57.1} & {53.1} & {51.1} & {94.6} & {47.3} & {98.2} & {66.9}\\[\methodsep]
& Ad-REM SVM & {51.6} & {53.1} & {46.8} & {86.3} & {54.7} & {92.8} & {64.2}\\
\end{tabular}
\end{table}

\begin{table}[ht]
 \caption{Accuracy on the Cross Dataset Testbed.}
 \label{tbl:accuracy-cdt}
 \tablesize
 \centering
\begin{tabular}{@{\hspace*{\leftsep}}ll@{\hspace*{\colsep}}c@{\hspace*{\colsep}}c@{\hspace*{\colsep}}c@{\hspace*{\colsep}}c@{\hspace*{\colsep}}c@{\hspace*{\colsep}}c@{\hspace*{\colsep}}c}
\hlinetop
& & C$\to$I & C$\to$S & I$\to$C & I$\to$S & S$\to$C & S$\to$I & avg\\
\hlinemid
\multicolumn{7}{l}{DECAF-fc7 features}\\[\subheadersep]
& Source SVM & {65.1} & {21.4} & {74.6} & {23.2} & {27.5} & {29.9} & {40.3}\\
& Source LR & {64.1} & {21.3} & {74.6} & {23.7} & {24.6} & {27.6} & {39.3}\\[\methodsep]
& SA & {43.7} & {13.9} & {52.0} & {15.1} & {15.8} & {14.3} & {25.8}\\
& FLDA-Q & {65.5} & {21.9} & {74.8} & {24.6} & {26.4} & {28.0} & {40.2}\\
& CORAL & {66.2} & {22.9} & {74.7} & {25.4} & {26.9} & {25.2} & {40.2}\\[\methodsep]
& Ad-REM SVM & {65.8} & {20.4} & {75.9} & {25.1} & {35.1} & {39.1} & {43.6}\\
\end{tabular}
\end{table}

\clearpage
\bibliography{da}

\begin{thebibliography}{46}
\providecommand{\natexlab}[1]{#1}
\providecommand{\url}[1]{\texttt{#1}}
\expandafter\ifx\csname urlstyle\endcsname\relax
  \providecommand{\doi}[1]{doi: #1}\else
  \providecommand{\doi}{doi: \begingroup \urlstyle{rm}\Url}\fi

\bibitem[Amini and Gallinari(2003)]{amini2003semi}
Massih-Reza Amini and Patrick Gallinari.
\newblock Semi-supervised learning with an explicit label-error model for
  misclassified data.
\newblock In \emph{Proceedings of the 18th IJCAI}, pages 555--560, 2003.

\bibitem[Ben-David et~al.(2007)Ben-David, Blitzer, Crammer, Pereira,
  et~al.]{ben2007analysis}
Shai Ben-David, John Blitzer, Koby Crammer, Fernando Pereira, et~al.
\newblock Analysis of representations for domain adaptation.
\newblock \emph{Advances in neural information processing systems},
  19:\penalty0 137, 2007.

\bibitem[Ben-David et~al.(2010)Ben-David, Blitzer, Crammer, Kulesza, Pereira,
  and Vaughan]{ben2010theory}
Shai Ben-David, John Blitzer, Koby Crammer, Alex Kulesza, Fernando Pereira, and
  Jennifer~Wortman Vaughan.
\newblock A theory of learning from different domains.
\newblock \emph{Machine learning}, 79\penalty0 (1-2):\penalty0 151--175, 2010.

\bibitem[Blitzer et~al.(2006)Blitzer, McDonald, and Pereira]{blitzer2006domain}
John Blitzer, Ryan McDonald, and Fernando Pereira.
\newblock Domain adaptation with structural correspondence learning.
\newblock In \emph{Proceedings of the 2006 conference on empirical methods in
  natural language processing}, pages 120--128. Association for Computational
  Linguistics, 2006.

\bibitem[Blitzer et~al.(2007)Blitzer, Dredze, Pereira,
  et~al.]{blitzer2007biographies}
John Blitzer, Mark Dredze, Fernando Pereira, et~al.
\newblock Biographies, bollywood, boom-boxes and blenders: Domain adaptation
  for sentiment classification.
\newblock In \emph{ACL}, volume~7, pages 440--447, 2007.

\bibitem[Bousmalis et~al.(2016)Bousmalis, Trigeorgis, Silberman, Krishnan, and
  Erhan]{bousmalis2016domain}
Konstantinos Bousmalis, George Trigeorgis, Nathan Silberman, Dilip Krishnan,
  and Dumitru Erhan.
\newblock Domain separation networks.
\newblock In \emph{Advances in Neural Information Processing Systems}, pages
  343--351, 2016.

\bibitem[Bruzzone and Marconcini(2010)]{bruzzone2010domain}
Lorenzo Bruzzone and Mattia Marconcini.
\newblock Domain adaptation problems: A dasvm classification technique and a
  circular validation strategy.
\newblock \emph{Pattern Analysis and Machine Intelligence, IEEE Transactions
  on}, 32\penalty0 (5):\penalty0 770--787, 2010.

\bibitem[Chollet et~al.(2015)]{chollet2015keras}
Fran\c{c}ois Chollet et~al.
\newblock Keras.
\newblock \url{https://github.com/fchollet/keras}, 2015.

\bibitem[Chopra et~al.(2013)Chopra, Balakrishnan, and Gopalan]{chopra2013dlid}
Sumit Chopra, Suhrid Balakrishnan, and Raghuraman Gopalan.
\newblock Dlid: Deep learning for domain adaptation by interpolating between
  domains.
\newblock In \emph{ICML workshop on challenges in representation learning},
  volume~2, 2013.

\bibitem[Fan et~al.(2008)Fan, Chang, Hsieh, Wang, and Lin]{Fan2008liblinear}
Rong-En Fan, Kai-Wei Chang, Cho-Jui Hsieh, Xiang-Rui Wang, and Chih-Jen Lin.
\newblock Liblinear: A library for large linear classification.
\newblock \emph{J. Mach. Learn. Res.}, 9:\penalty0 1871--1874, June 2008.
\newblock ISSN 1532-4435.

\bibitem[Fernando et~al.(2013)Fernando, Habrard, Sebban, and
  Tuytelaars]{Fernando2013SA}
Basura Fernando, Amaury Habrard, Marc Sebban, and Tinne Tuytelaars.
\newblock Unsupervised visual domain adaptation using subspace alignment.
\newblock In \emph{Proceedings of the 2013 IEEE International Conference on
  Computer Vision}, ICCV '13, pages 2960--2967, Washington, DC, USA, 2013. IEEE
  Computer Society.
\newblock ISBN 978-1-4799-2840-8.

\bibitem[Franc et~al.(2011)Franc, Zien, and Sch{\"o}lkopf]{franc2011support}
Vojtech Franc, Alexander Zien, and Bernhard Sch{\"o}lkopf.
\newblock Support vector machines as probabilistic models.
\newblock In \emph{Proceedings of the 28th International Conference on Machine
  Learning (ICML-11)}, pages 665--672, 2011.

\bibitem[Ganin and Lempitsky(2015)]{ganin2015unsupervised}
Yaroslav Ganin and Victor Lempitsky.
\newblock Unsupervised domain adaptation by backpropagation.
\newblock In \emph{International Conference on Machine Learning}, pages
  1180--1189, 2015.

\bibitem[Ganin et~al.(2016)Ganin, Ustinova, Ajakan, Germain, Larochelle,
  Laviolette, Marchand, and Lempitsky]{ganin2016domain}
Yaroslav Ganin, Evgeniya Ustinova, Hana Ajakan, Pascal Germain, Hugo
  Larochelle, Fran{\c{c}}ois Laviolette, Mario Marchand, and Victor Lempitsky.
\newblock Domain-adversarial training of neural networks.
\newblock \emph{Journal of Machine Learning Research}, 17\penalty0
  (59):\penalty0 1--35, 2016.

\bibitem[Germain et~al.(2016)Germain, Habrard, Laviolette, and
  Morvant]{germain2016new}
Pascal Germain, Amaury Habrard, Fran{\c{c}}ois Laviolette, and Emilie Morvant.
\newblock A new {PAC}-{B}ayesian perspective on domain adaptation.
\newblock In \emph{International Conference on Machine Learning (ICML)}, 2016.

\bibitem[Ghahramani and Jordan(1994)]{ghahramani1994supervised}
Zoubin Ghahramani and Michael~I Jordan.
\newblock Supervised learning from incomplete data via an em approach.
\newblock In \emph{Advances in neural information processing systems}, pages
  120--127, 1994.

\bibitem[Gong et~al.(2012)Gong, Shi, Sha, and Grauman]{gong2012geodesic}
Boqing Gong, Yuan Shi, Fei Sha, and Kristen Grauman.
\newblock Geodesic flow kernel for unsupervised domain adaptation.
\newblock In \emph{Computer Vision and Pattern Recognition (CVPR), 2012 IEEE
  Conference on}, pages 2066--2073. IEEE, 2012.

\bibitem[Gong et~al.(2013)Gong, Grauman, and Sha]{gong2013connecting}
Boqing Gong, Kristen Grauman, and Fei Sha.
\newblock Connecting the dots with landmarks: Discriminatively learning
  domain-invariant features for unsupervised domain adaptation.
\newblock In \emph{ICML (1)}, pages 222--230, 2013.

\bibitem[Grandvalet and Bengio(2006)]{grandvalet2006entropy}
Yves Grandvalet and Yoshua Bengio.
\newblock Entropy regularization.
\newblock \emph{Semi-supervised learning}, pages 151--168, 2006.

\bibitem[Grandvalet et~al.(2006)Grandvalet, Mari{\'e}thoz, and
  Bengio]{grandvalet2006probabilistic}
Yves Grandvalet, Johnny Mari{\'e}thoz, and Samy Bengio.
\newblock A probabilistic interpretation of svms with an application to
  unbalanced classification.
\newblock In \emph{Advances in Neural Information Processing Systems}, pages
  467--474, 2006.

\bibitem[Habrard et~al.(2013)Habrard, Peyrache, and
  Sebban]{habrard2013boosting}
Amaury Habrard, Jean-Philippe Peyrache, and Marc Sebban.
\newblock Boosting for unsupervised domain adaptation.
\newblock In \emph{Joint European Conference on Machine Learning and Knowledge
  Discovery in Databases}, pages 433--448. Springer, 2013.

\bibitem[He et~al.(2016)He, Zhang, Ren, and Sun]{he2016deep}
Kaiming He, Xiangyu Zhang, Shaoqing Ren, and Jian Sun.
\newblock Deep residual learning for image recognition.
\newblock In \emph{Proceedings of the IEEE conference on computer vision and
  pattern recognition}, pages 770--778, 2016.

\bibitem[Joachims(2006)]{joachims2006transductive}
Thorsten Joachims.
\newblock Transductive support vector machines.
\newblock \emph{Chapelle et al.(2006)}, pages 105--118, 2006.

\bibitem[Kouw et~al.(2016)Kouw, van~der Maaten, Krijthe, and
  Loog]{kouw2016feature}
Wouter~M Kouw, Laurens~JP van~der Maaten, Jesse~H Krijthe, and Marco Loog.
\newblock Feature-level domain adaptation.
\newblock \emph{Journal of Machine Learning Research}, 17\penalty0
  (171):\penalty0 1--32, 2016.

\bibitem[Li et~al.(2008)Li, Guan, Li, and Chin]{li2008self}
Yuanqing Li, Cuntai Guan, Huiqi Li, and Zhengyang Chin.
\newblock A self-training semi-supervised svm algorithm and its application in
  an eeg-based brain computer interface speller system.
\newblock \emph{Pattern Recognition Letters}, 29\penalty0 (9):\penalty0
  1285--1294, 2008.

\bibitem[Long et~al.(2015)Long, Cao, Wang, and Jordan]{long2015learning}
Mingsheng Long, Yue Cao, Jianmin Wang, and Michael Jordan.
\newblock Learning transferable features with deep adaptation networks.
\newblock In \emph{Proceedings of The 32nd International Conference on Machine
  Learning}, pages 97--105, 2015.

\bibitem[Long et~al.(2016{\natexlab{a}})Long, Wang, and Jordan]{long2016deep}
Mingsheng Long, Jianmin Wang, and Michael~I Jordan.
\newblock Deep transfer learning with joint adaptation networks.
\newblock \emph{arXiv preprint arXiv:1605.06636}, 2016{\natexlab{a}}.

\bibitem[Long et~al.(2016{\natexlab{b}})Long, Wang, and
  Jordan]{long2016unsupervised}
Mingsheng Long, Jianmin Wang, and Michael~I Jordan.
\newblock Unsupervised domain adaptation with residual transfer networks.
\newblock \emph{arXiv preprint arXiv:1602.04433}, 2016{\natexlab{b}}.

\bibitem[Margolis(2011)]{margolis2011literature}
Anna Margolis.
\newblock A literature review of domain adaptation with unlabeled data.
\newblock \emph{Tec. Report}, pages 1--42, 2011.

\bibitem[McLachlan(1975)]{mclachlan1975iterative}
Geoffrey~J McLachlan.
\newblock Iterative reclassification procedure for constructing an
  asymptotically optimal rule of allocation in discriminant analysis.
\newblock \emph{Journal of the American Statistical Association}, 70\penalty0
  (350):\penalty0 365--369, 1975.

\bibitem[Nigam et~al.(2006)Nigam, McCallum, and Mitchell]{nigam2006semi}
Kamal Nigam, Andrew McCallum, and Tom Mitchell.
\newblock Semi-supervised text classification using em.
\newblock \emph{Semi-Supervised Learning}, pages 33--56, 2006.

\bibitem[Pan and Yang(2010)]{pan2010survey}
Sinno~Jialin Pan and Qiang Yang.
\newblock A survey on transfer learning.
\newblock \emph{IEEE Transactions on knowledge and data engineering},
  22\penalty0 (10):\penalty0 1345--1359, 2010.

\bibitem[Patel et~al.(2015)Patel, Gopalan, Li, and Chellappa]{patel2015visual}
Vishal~M Patel, Raghuraman Gopalan, Ruonan Li, and Rama Chellappa.
\newblock Visual domain adaptation: A survey of recent advances.
\newblock \emph{IEEE signal processing magazine}, 32\penalty0 (3):\penalty0
  53--69, 2015.

\bibitem[Saenko et~al.(2010)Saenko, Kulis, Fritz, and
  Darrell]{saenko2010adapting}
Kate Saenko, Brian Kulis, Mario Fritz, and Trevor Darrell.
\newblock Adapting visual category models to new domains.
\newblock In \emph{European conference on computer vision}, pages 213--226.
  Springer, 2010.

\bibitem[Seeger(2000)]{seeger2000learning}
Matthias Seeger.
\newblock Learning with labeled and unlabeled data.
\newblock Technical report, 2000.

\bibitem[Sener et~al.(2016)Sener, Song, Saxena, and
  Savarese]{sener2016learning}
Ozan Sener, Hyun~Oh Song, Ashutosh Saxena, and Silvio Savarese.
\newblock Learning transferrable representations for unsupervised domain
  adaptation.
\newblock In \emph{Advances in Neural Information Processing Systems}, pages
  2110--2118, 2016.

\bibitem[Shimodaira(2000)]{shimodaira2000improving}
Hidetoshi Shimodaira.
\newblock Improving predictive inference under covariate shift by weighting the
  log-likelihood function.
\newblock \emph{Journal of statistical planning and inference}, 90\penalty0
  (2):\penalty0 227--244, 2000.

\bibitem[Simonyan and Zisserman(2014)]{Simonyan14vgg}
Karen Simonyan and Andrew Zisserman.
\newblock Very deep convolutional networks for large-scale image recognition.
\newblock \emph{CoRR}, abs/1409.1556, 2014.

\bibitem[Smola et~al.(2005)Smola, Vishwanathan, and Hofmann]{smola2005kernel}
Alexander~J Smola, SVN Vishwanathan, and Thomas Hofmann.
\newblock Kernel methods for missing variables.
\newblock In \emph{AISTATS}, 2005.

\bibitem[Sun and Saenko(2016)]{sun2016deep}
Baochen Sun and Kate Saenko.
\newblock Deep coral: Correlation alignment for deep domain adaptation.
\newblock In \emph{Computer Vision--ECCV 2016 Workshops}, pages 443--450.
  Springer, 2016.

\bibitem[Sun et~al.(2016)Sun, Feng, and Saenko]{sun2016return}
Baochen Sun, Jiashi Feng, and Kate Saenko.
\newblock Return of frustratingly easy domain adaptation.
\newblock In \emph{Thirtieth AAAI Conference on Artificial Intelligence}, 2016.

\bibitem[Tan et~al.(2009)Tan, Cheng, Wang, and Xu]{tan2009adapting}
Songbo Tan, Xueqi Cheng, Yuefen Wang, and Hongbo Xu.
\newblock Adapting naive bayes to domain adaptation for sentiment analysis.
\newblock \emph{Advances in Information Retrieval}, pages 337--349, 2009.

\bibitem[Tommasi and Tuytelaars(2014)]{tommasi2014testbed}
Tatiana Tommasi and Tinne Tuytelaars.
\newblock A testbed for cross-dataset analysis.
\newblock In \emph{European Conference on Computer Vision}, pages 18--31.
  Springer, 2014.

\bibitem[Tzeng et~al.(2014)Tzeng, Hoffman, Zhang, Saenko, and
  Darrell]{tzeng2014deep}
Eric Tzeng, Judy Hoffman, Ning Zhang, Kate Saenko, and Trevor Darrell.
\newblock Deep domain confusion: Maximizing for domain invariance.
\newblock \emph{arXiv preprint arXiv:1412.3474}, 2014.

\bibitem[Tzeng et~al.(2015)Tzeng, Hoffman, Darrell, and
  Saenko]{tzeng2015simultaneous}
Eric Tzeng, Judy Hoffman, Trevor Darrell, and Kate Saenko.
\newblock Simultaneous deep transfer across domains and tasks.
\newblock In \emph{Proceedings of the IEEE International Conference on Computer
  Vision}, pages 4068--4076, 2015.

\bibitem[Zhang et~al.(2015)Zhang, Yu, Chang, and Wang]{zhang2015deep}
Xu~Zhang, Felix~Xinnan Yu, Shih-Fu Chang, and Shengjin Wang.
\newblock Deep transfer network: Unsupervised domain adaptation.
\newblock \emph{arXiv preprint arXiv:1503.00591}, 2015.

\end{thebibliography}

\end{document}